\documentclass[11pt,a4paper]{article}
\usepackage[hyperref]{eacl2021}
\usepackage{times}
\usepackage{latexsym}

\usepackage[ruled,vlined]{algorithm2e}
\usepackage{multirow,multicol}

\usepackage{microtype}
\usepackage{graphicx}
\usepackage{lipsum}
\usepackage{booktabs}

\aclfinalcopy %

\title{Hate-Alert@DravidianLangTech-EACL2021: Ensembling strategies for Transformer-based Offensive language Detection}

\author {
        Debjoy Saha ,
        Naman Paharia ,
        Debajit Chakraborty,\\
        \textbf{Punyajoy Saha}, 
        \textbf{Animesh Mukherjee}\\
        Indian Institute of Technology, Kharagpur, India \\
        sahadebjoy10@iitkgp.ac.in, namanpaharia.27@gmail.com, debajit15@iitkgp.ac.in\\ punyajoys@iitkgp.ac.in, animeshm@cse.iitkgp.ac.in
}

\date{}

\begin{document}
\maketitle
\begin{abstract}
Social media often acts as breeding grounds for different forms of offensive content. For low resource languages like Tamil, the situation is more complex due to the poor performance of multilingual or language-specific models and lack of proper benchmark datasets. Based on this shared task ``Offensive Language Identification in Dravidian Languages'' at EACL 2021, we present an exhaustive exploration of different transformer models, We also provide a genetic algorithm technique for ensembling different models. Our ensembled models trained separately for each language secured the \textbf{first} position in Tamil, the \textbf{second} position in Kannada, and the \textbf{first} position in Malayalam sub-tasks. The models and codes are provided\footnote{https://github.com/Debjoy10/Hate-Alert-DravidianLangTech}.
\end{abstract}

\section{Introduction}

Social media platforms have become a prominent way of communication, be it for acquiring information or promotion of
business\footnote{https://www.webfx.com/internet-marketing/social-media-marketing-advantages-and-disadvantages.html}. While we cannot deny the positives, there are some ill consequences of social media as well. Bad actors often use different social media platforms by posting tweets/comments that insult others by targeting their culture and beliefs. In social media, such posts are collectively known as offensive language~\cite{chen2012detecting}. To reduce offensive content, different social media platforms like YouTube have laid down moderation policies and employ moderators for maintaining civility in their platforms. Recently, the moderators are finding it difficult~\cite{Googlean72:online} to continue the moderation due to the ever-increasing volume of offensive data. Hence, platforms are looking toward automatic moderation systems. For instance, Facebook~\cite{Facebook34:online} is proactively removing a large part of the harmful content from its platform, even before the users report them. 
There are concerns by different policy-makers that these automatic moderation systems may be erroneous\footnote{https://www.forbes.com/sites/johnkoetsier/2020/06/09/300000-facebook-content-moderation-mistakes-daily-report-says/}. Situation for countries like India is more complex, as courts often face dilemma while interpreting harmful content and social platforms like Facebook are often unable to take necessary actions\footnote{https://www.npr.org/2020/11/27/939532326/facebook-accused-of-violating-its-hate-speech-policy-in-india}. Hence, more effort is required to detect and mitigate offensive language in the Indian social media. 

Recently, different shared tasks like HASOC 2019\footnote{https://hasocfire.github.io/hasoc/2019/index.html} have been launched to understand hateful and offensive language in Indian context but it is limited to Hindi and English mostly. A sub-task in HASOC 2020\footnote{https://sites.google.com/view/dravidian-codemix-fire2020/overview} aimed to detect offensive posts in a code-mixed dataset. Extending that task further, the organisers of this shared task have put together a large dataset of 43919, 7772, 20010 posts in three Dravidian languages -- Tamil, Kannada, Malayalam respectively, to further advance research on offensive posts in these languages. In this paper, we aim to build algorithmic systems that can detect offensive posts. Contributions of our paper are two-fold. First, we investigate how the current state-of-the-art multilingual language models perform on these languages. Second, we demonstrate how we can use ensembling techniques to improve our classification performance.

\begin{table*}[t]
\centering
\footnotesize{
\begin{tabular}{{llllllllllllllllll}}
\toprule 
    Classifiers & \multicolumn{3}{c}{Tamil} & \multicolumn{3}{c}{Kannada} & \multicolumn{3}{c}{Malayalam}\\
    \midrule
    & Train & Dev & Test
    & Train & Dev & Test
    & Train & Dev & Test\\
    \midrule
    Not-offensive &25425    & 3193    &3190  &3544 &426 &427 &14153 &1779 &1765 \\
    Offensive-untargeted &2906    &356    &368  &212 &33 &33 &191 &20 &29 \\
    Offensive-targeted-individual &2343    &307    &315  &487 &66 &75 &239 &24 &27 \\
    Offensive-targeted-group &2557    &295    &288  &329 &45 &44 &140 &13 &23 \\
    Offensive-targeted-other &454    &65    &71  &123 &16 &14 &- &- &- \\
    Not-in-indented-language &1454    &172    &160  &1522 &191 &185 &1287 &163 &157 \\
    Total &35139    &4388    &4392  &6217 &777 &778 &16010 &1999
    &2001 \\
    \bottomrule
\end{tabular}
}
\caption{Dataset statistics for languages Tamil, Kannada and Malayalam for all splits Train, Dev and Test}
\label{table:Data}
\end{table*}

\section{Related Work}

Offensive language has been studied in the research community for a long time, One of the earliest studies~\cite{6406271} tried to detect offensive users by using lexical syntactic features generated from their posts. Although, they provided an efficient framework for future research, their dataset was small for any conclusive evidence. \citeauthor{davidson2017automated} curated one of the largest dataset containing both offensive and hate speech. The authors found that one of the issues with their best performing models was that they could not distinguish between hate and offensive posts. In order to mitigate this, subsequent research~\cite{pitsilis2018detecting} tried to use deep learning to identify offensive language in English and found that recurrent neural networks (RNNs) are quite effective this task. Recently, the research community has begun to focus on offensive language detection in other low resourced languages like Danish~\cite{sigurbergsson2019offensive}, Greek~\cite{pitenis2020offensive} and Turkish~\cite{ccoltekin2020corpus}. In the Indian context, the HASOC 2019 shared task~\cite{mandl2019overview} was a significant effort in that direction, where the authors developed a dataset of hate and offensive posts in Hindi and English. The best model in this competition used an ensemble of multilingual transformers, fine-tuned on the given dataset~\cite{mishra20193idiots}. In Dravidian part of HASOC 2020, ~\citeauthor{renjit2020cusatnlp} used an ensemble of deep learning and simple neural networks to identify offensive posts in Manglish (Malayalam in roman font).

Transformer based language models are becoming quite popular in the past few years. Recently, different multilingual models like XLM-RoBERTa \citep{conneau2019unsupervised}, multilingual-BERT~\cite{devlin2018bert}, MuRIL~\footnote{https://tfhub.dev/google/MuRIL/1} and Indic-BERT~\cite{conneau2019unsupervised} have been introduced to facilitate NLP research in different languages. Often in different machine learning pipeline, ensembling different classification outcomes helps in getting better performance~\cite{alonso2020hate,renjit2020cusatnlp,mishra20193idiots}. Rather than selecting the models for the ensemble manually, genetic algorithms (GA) are used to optimise the weights of different classifiers, to improve the ensemble performance on the development set. GA-based ensembling techniques have previously been used in the hate speech domain for architecture and hyperparamter search~\citet{9308419}. 

\section{Dataset description}
The shared task on Offensive Language Identification in Dravidian Languages-EACL 2021 \citep{dravidianoffensive-eacl} is based on a post classification problem with an aim to moderate and minimise offensive content in social media. The objective of the shared task is to develop methodology and language models for code-mixed data in low-resource languages, as models trained on monolingual data fail to comprehend the semantic complexity of a code-mixed dataset. 

\paragraph{Dataset:} The Dravidian offensive code-mixed language dataset is available for Tamil \citep{chakravarthi-etal-2020-corpus}, Kannada \citep{hande-etal-2020-kancmd} and Malayalam \citep{chakravarthi-etal-2020-sentiment}. The data provided is scraped entirely from the YouTube comments of a multilingual community where code-mixing is a prevalent phenomenon. The dataset contains rows of text and the corresponding labels from the list not-offensive, offensive-untargeted, offensive-targeted-individual, offensive-targeted-group, offensive-targeted-other, or not-in-indented-language. Final evaluation score was calculated using weighted F1-score metric on a held-out test dataset.

We present the dataset statistics in Table \ref{table:Data}. Please note that the Malayalam split of the dataset contained no instances of 'Offensive-targeted-other' label, so classification is done using 5 labels only, instead of the original six labels. In order to understand the amount of misspelt and code-mixed words, we compare with an existing pure language vocabulary available in the Dakshina dataset~\citep{roark2020processing}. We find the proportion of out-of-vocabulary (OOV) words (including code-mixed, English and misspelt words) in the dataset as 85.55{\%}, 84.23{\%} and 83.03{\%} in Tamil, Malayalam and Kannada respectively.

\section{Methodology}
In this section, we discuss the different parts of the pipeline that we followed to detect offensive posts in this dataset.
\subsection{Machine learning models}
\label{baseline}
As a part of our initial experiments, we used several machine learning models to establish a baseline performance. We employed random forests, logistic regression and trained them with TF-IDF vectors. The best results were obtained on ExtraTrees Classifier~\citep{10.1007/s10994-006-6226-1} with 0.70, 0.63 and 0.95 weighted F1-scores on Tamil, Kannada and Malayalam respectively. As we will notice further, these performances were lower than single transformer based model. Hence, the simple machine learning models were not used in the subsequent analysis.
 
\subsection{Transformer models}
\label{trnbasic}
One of the issues with simple machine learning models is the inability to learn the context of a word based on its neighbourhood. Recent transformer based architectures are capable of capturing this context, as established by their superior performance in different downstream tasks. For our purpose, we fine-tuned different state-of-the-art multilingual BERT models on the given datasets. This includes XLM-RoBERTa~\cite{conneau2019unsupervised}, multilingual-BERT~\cite{devlin2018bert}\footnote{XLM-Roberta-Base, 270M parameters, trained on data from 100 languages; Multilingual-BERT-Base, 179M parameters, trained on data from the top 104 languages.}, Indic BERT and MuRIL\footnote{Originally released by Google, MuRIL (Multilingual Representations for Indian Languages) is a BERT model pre-trained on code-mixed data from 17 Indian languages \url{https://huggingface.co/simran-kh/muril-cased-temp}}. We also pretrain XLM-Roberta-Base on the target dataset for 20 epochs using Masked Language Modeling, to capture the semantics of the code-mixed corpus. This additional pretrained BERT model was also used for fine-tuning. In addition, all models were fine-tuned separately using unweighted and weighted cross-entropy loss functions~\cite{10.1145/1102351.1102422}. 
For training, we use HuggingFace~\cite{wolf2019huggingface} with PyTorch~\cite{paszke2019pytorch}. We use the Adam adaptive optimizer~\cite{loshchilov2019decoupled} with an initial learning rate of 1e-5. Training is stopped by early stopping if macro-F1 score of the development split of the dataset does not increase for 5 epochs.

\begin{table}[t]
\centering
\footnotesize{
\begin{tabular}{{llllllllllllllllll}}
\toprule 
    Classifiers & \multicolumn{2}{c}{Tamil} & \multicolumn{2}{c}{Kannada} & \multicolumn{2}{c}{Malayalam} \\
    \midrule
    & Dev & Test
    & Dev & Test
    & Dev & Test \\
    \midrule
    XLMR-base (A) &0.77    &0.76    &0.69  &0.70 &0.97 &0.96 \\
    XLMR-large &\textbf{0.78}    &\textbf{0.77}   &0.69  &0.71 &\textbf{0.97} &\textbf{0.97} \\
    XLMR-C (B) &0.76    &0.76    &\textbf{0.70}  &\textbf{0.73} &\textbf{0.97} &\textbf{0.97} \\
    mBERT-base (C) &0.73    &0.72    &0.69  &0.70 &0.97 &0.96 \\
    IndicBERT &0.73    &0.71    &0.62  &0.66 &0.96 &0.95 \\
    MuRIL &0.75    &0.74    &0.67  &0.67 &0.96 &0.96\\
    DistilBERT &0.74    &0.74    &0.68  &0.69 &0.96 &0.95 \\
    \midrule
    CNN &0.71 &0.70    &0.60  &0.61 &0.95 &0.95\\
    \midrule
    CNN + A + C &0.78    &0.76 &0.71  &0.70 &\textbf{0.97} &\textbf{0.97} \\
    CNN + A + B &\textbf{0.78}    &\textbf{0.77}    &0.71  &0.71 &\textbf{0.97} &\textbf{0.97} \\
    CNN + B + C &0.77    &0.76    &\textbf{0.71}  &\textbf{0.72} &\textbf{0.97} &\textbf{0.97} \\
    \bottomrule
\end{tabular}
}
\caption{Weighted F1-score comparison for transformer, CNN and fusion models on Dev and Test splits (XLMR-C refers to the custom-pretrained XLM-Roberta-Base Classifier).}
\label{table:Transformer}
\end{table}

\begin{figure}
 \centering
 \includegraphics[width=0.7\columnwidth]{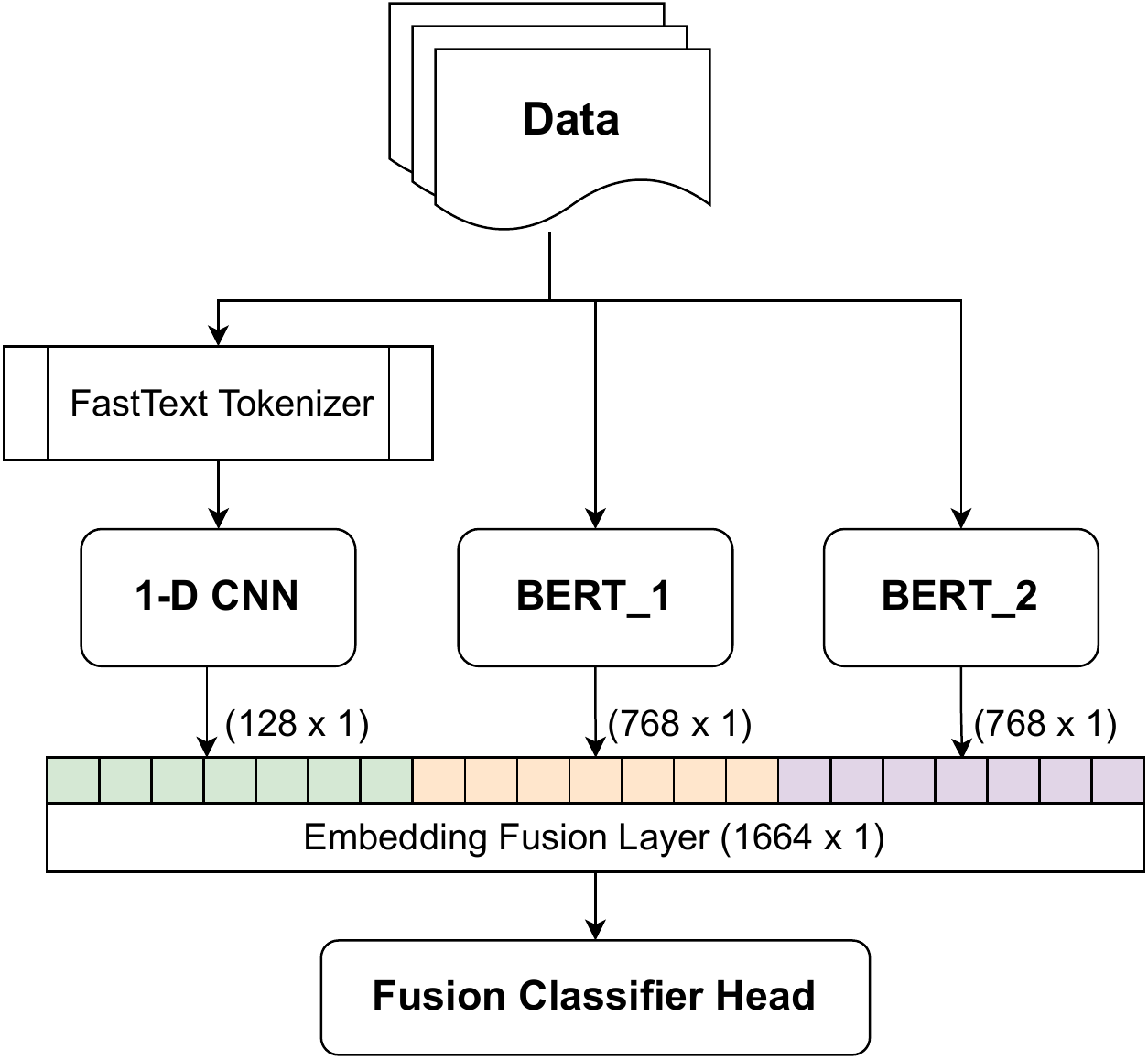}
 \caption{Our fusion model architecture for two BERT models. Note that $768 \times 1$ embedding sizes are used for the BERT-base models. Embeddings size of $1024 \times 1$ is used for BERT-large models.}
 \label{fig:pipeline}
\end{figure}

\subsection{Fusion models}

Convolution neural networks are able to capture neighbourhood information more effectively. One of the previous state-of-the-art model to detect hate speech was CNN-GRU~\cite{cnn-gru}, We propose a new BERT-CNN fusion classifier where we train a single classification head on the concatenated embeddings from different BERT and CNN models. BERT models were initialised with the fine-tuned weights in the former section and the weights were frozen. The number of BERT models in a single fusion model was kept flexible with maximum number of models fixed to three, due to memory limitation. For the CNN part, we use the 128-dim final layer embeddings from CNN models trained on skip-gram word vectors using FastText~\citep{bojanowski2017enriching}\footnote{\url{https://fasttext.cc/docs/en/unsupervised-tutorial.html}}. FastText vectors worked the best among other word embeddings like LASER~\citep{artetxe2019massively}. For the fusion classifier head, we use a feed-forward neural network having four layers with batch normalization~\citep{ioffe2015batch} and dropout~\citep{JMLR:v15:srivastava14a} on the final layer. The predictions were generated from a softmax layer of dimension equal to the number of classes. We present the details of the pipeline in Figure \ref{fig:pipeline}.

\begin{table}[t]
\centering
\footnotesize{
\begin{tabular}{lllllll}
\toprule 
    Model Sets & \multicolumn{2}{c}{Tamil} & \multicolumn{2}{c}{Kannada} & \multicolumn{2}{c}{Malayalam}\\
    \midrule
    &  Dev & Test
    &  Dev & Test
    &  Dev & Test \\
    \midrule
    Transformers &0.80 & 0.78&0.74&0.73&0.98&0.97\\
    F-models  & 0.79&0.77&0.73&0.73&0.98&0.97\\
    R-models &0.79&0.78&0.75&0.74&0.97&0.97 \\
    \midrule
    \textbf{Overall} &\textbf{0.80}&\textbf{0.78}&\textbf{0.75}&\textbf{0.74}&\textbf{0.98}&\textbf{0.97} \\
    \bottomrule
\end{tabular}
}
\caption{Weighted-F1 score comparison for GA-weighted ensemble for transformers category, Fusion models(F-models) and Random seed models(R-models)}
\label{table:Ensemble}
\end{table}

\subsection{Ensembling strategies}
\label{Ensemble}
Ensemble of different models often turn out better predictors than using a single classifier. Standard prediction averaging ensembles will not perform well, since some models might be weak predictors in the mix of different models. One of the strategies to reduce the influence of weak models is using weights for different models based on their performance. Genetic algorithm (GA) based techniques~\cite{9308419} are one of the popular ways to set the weights of different models in an ensemble. Our approach is similar to that introduced in \citet{10.5555/1642194.1642200}, except that instead of selecting the models with the highest weights for the final ensemble, we directly use the weights to compute the weighted average ensemble. 

Another issue with neural networks is the performance is dependent on the initial random seeds. With pretrained models like BERT, most of the weights are fixed only in the final layer (classification head). Past research~\cite{mccoy2020berts} has shown that even the initialisation of this final layer can affect the final performance by large margins. Hence, we take 10 different random seeds to train the models and then pass all the models to the GA pipeline. We perform this operation for two of the best models in Table \ref{table:Transformer}.

\section{Results and conclusion}

We observe that among the individual transformer models, the best performance is obtained using XLM-RoBERTa-large (XLMR-large) in the Tamil dataset and Custom XLM-RoBERTa-base (XLMR-C) in the Kannada dataset. For Malayalam dataset, both the former models perform similarly. The higher performance of XLM-RoBERTa~\cite{artetxe2019massively} models can be attributed to the fact that they are pretrained using a parallel corpus (same corpus in different languages). Further pretraining with our dataset helps in further improvement of the performance in the Kannada dataset. We did not use the XLM-R large model further due to limited GPU space. Next, we note the performance of the fusion models, which perform almost similarly across different combinations. 

When we use different random seeds, the performance of multilingual BERT models varied around 2-3\% across different languages. For XLM-RoBERTa models the variation was more (around 15-20\%). Table \ref{table:Ensemble} shows the ensemble performance of different categories of models and all the models combined. GA-optimised weighted ensembling improves the final model scores by a 1-2\% across datasets of different languages which finally helped us to rank higher in the leader board.

In this shared task, we evaluated different transformer based architectures and introduced different ensembling strategies. We found that XLM-RoBERTa models usually perform better than other transformer models, although their performance is highly variable across different random seeds. GA based ensembling helps us in further improving the models. Our immediate next step will be to investigate the reason behind lower performance of IndicBERT and MuRIL which are specifically trained for Indian context.

\bibliography{anthology,eacl2021}
\bibliographystyle{acl_natbib}

\end{document}